\documentclass[journal]{IEEEtran}

\usepackage{graphics}
\usepackage{epsfig}
\usepackage{mathptmx}
\usepackage{times}
\usepackage{amsmath}
\usepackage{amssymb}
\usepackage{caption}
\usepackage{multirow}
\usepackage{hyperref}
\usepackage[]{xcolor}
\usepackage{booktabs}
\usepackage{algorithm}
\usepackage{algorithmic}
\usepackage{pifont}

\title{TAGA: A Tangent-Based Reactive Approach for Socially Compliant Robot Navigation Around Human Groups}

\author{Utsha Kumar Roy$^{1}$ and Sejuti Rahman$^{2}$%
\thanks{Code is available upon request via email to the corresponding author.}%
\thanks{$^{1}$Utsha Kumar Roy is with the Department of Computer Science and Engineering,
        BRAC University, Dhaka, Bangladesh
        {\tt\small utsha.roy@bracu.ac.bd}}%
\thanks{$^{2}$Sejuti Rahman is with the New Uzbekistan University,
        Tashkent, Uzbekistan
        {\tt\small s.rahman@newuu.uz}}%
}

\begin{document}

\maketitle
\thispagestyle{empty}
\pagestyle{empty}

\begin{abstract}
Robots navigating in human-populated environments must not only avoid collisions but also respect the social structure of human crowds, particularly the implicit boundaries of social groups. Most existing navigation approaches model humans as independent individuals, leading to socially disruptive behavior even when collision-free motion is achieved. This paper presents TAGA (Tangent Action for Group Avoidance), a modular and lightweight framework that explicitly accounts for human group formations during robot navigation. TAGA applies a tangent-based action strategy to guide the robot around detected group boundaries while maintaining efficient progress toward the goal, and requires no retraining of the underlying navigation policy. A hierarchical safety controller coordinates group-level avoidance with individual collision prevention. We further propose the Group Crossing Rate (GCR) as a continuous complementary metric that quantifies the fraction of timesteps the robot spends inside any group convex hull throughout the navigation episode, providing finer-grained assessment than terminal-state metrics alone. To support rigorous evaluation, we introduce a realistic crowd simulation benchmark incorporating five empirically motivated modeling phases: individual speed variation, dynamic group speed coupling, F-formations for static groups, leader--follower dynamics, and convex-hull group boundaries. Extensive experiments across reactive methods (ORCA, Social Force) and a learning-based method (Intention-RL) reveal a consistent \textit{reactive-learning asymmetry}: TAGA significantly improves social compliance and success rate for reactive baselines while degrading performance for learned policies that have already internalized group-routing behavior through training. These findings provide actionable guidance for when modular group-awareness adds the most value versus when end-to-end group-aware training is the better choice.

\end{abstract}

\section{INTRODUCTION}

Consider a service robot navigating a busy train station concourse or shopping mall. The crowd it encounters is not a stream of independent walkers moving in isolation — it is socially structured: families walking side-by-side, colleagues in conversation, tour groups proceeding in loose formation. Empirical pedestrian studies report that the majority of people in public spaces belong to a social group at any given moment~\cite{moussaid2010walking}, with surveys in commercial and transit spaces placing the fraction as high as 70\%. These groups are not merely clusters of obstacles; they maintain shared spatial conventions~\cite{kendon1990conducting} and proximity norms~\cite{hall1966hidden} that a well-behaved agent navigating alongside humans is expected to honor.

\begin{figure}[!t]
\centering
\includegraphics[width=\columnwidth]{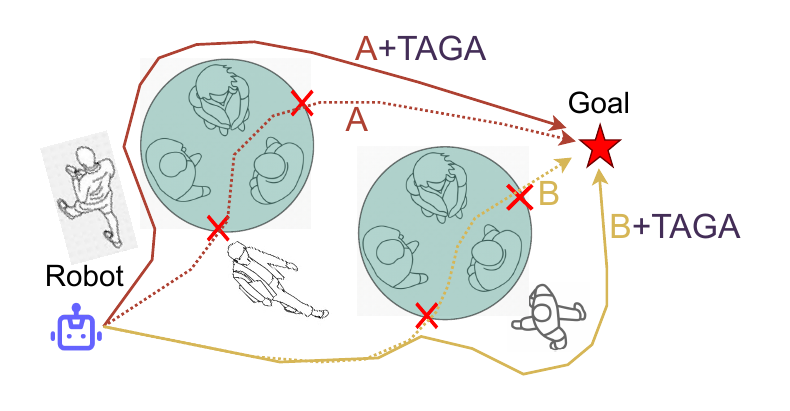}
\vspace{-20pt}
\caption{Baseline navigation methods A and B (dotted lines) cut through group spaces, violating social norms (marked \textcolor{red}{$\times$}). TAGA-enhanced paths (solid lines) guide the robot around group boundaries while efficiently reaching the goal \textcolor{red}{$\bigstar$}.}
\vspace{-14pt}
\label{fig-env-theme}
\end{figure}

When a robot passes through the interior of a social group — even without physical contact — it disrupts the spatial arrangement that the interaction depends on, forces members to scatter and regroup, and signals to bystanders a failure to treat the crowd as socially organized rather than as a field of obstacles. In shared public spaces, how a robot arrives matters alongside whether it arrives: a system that achieves high task success while routinely violating group boundaries will not earn the acceptance needed for sustainable real-world deployment~\cite{gao2022evaluation}.

A fundamental obstacle to progress in this area has been the gap between benchmark simulation and real crowd behavior. Standard evaluation environments model pedestrians as identical agents with a fixed preferred speed and no social grouping~\cite{chen2017decentralized, liu2020decentralized, liu2023intention}. Real crowds diverge from this in three empirically documented ways: individual walking speeds are heterogeneous~\cite{weidmann1992social}, groups reduce their collective pace to accommodate their slowest member~\cite{moussaid2010walking}, and stationary groups adopt sociologically studied spatial formations (F-formations~\cite{kendon1990conducting}) while moving groups exhibit leader-follower locomotion~\cite{helbing1995social}. Policies benchmarked on simplified environments may appear competent in simulation yet be unprepared for the richer structure of realistic crowds.

Existing group-aware navigation methods address some of these challenges but share notable limitations. Reinforcement learning policies with group-aware rewards~\cite{group-aware, lu2025garn} embed group awareness inside a monolithic trained policy, requiring full retraining to be applied to any new base policy or environment. SANG~\cite{schmuck2025sang} addresses only static conversational groups and has not been evaluated under the mixed static-dynamic group populations that characterize real public spaces. Moreover, none of these methods evaluates navigation under multiple pedestrian dynamics models, leaving open the question of whether good performance generalizes across crowd behavior assumptions or is specific to one training distribution. Separating a realistic benchmark from the methods evaluated on it is essential to answer this question.

This paper makes two peer-level contributions that address both the benchmarking and the methods gap.

\textbf{Contribution 1 — Realistic crowd simulation benchmark.}
We extend the simulation framework of Liu et al.~\cite{liu2023intention} with five empirically grounded modeling phases: individual speed heterogeneity drawn from Weidmann's pedestrian data~\cite{weidmann1992social}, group speed coupling~\cite{moussaid2010walking}, F-formation static groups~\cite{kendon1990conducting}, leader-follower moving groups~\cite{helbing1995social}, and convex-hull group boundaries. The benchmark supports two independent pedestrian dynamics models (ORCA~\cite{orca} and Social Force~\cite{helbing1995social}) so that generalization across crowd behavior assumptions can be assessed. We also introduce the \textbf{Group Crossing Rate (GCR)}, a continuous non-terminal metric that quantifies the fraction of navigation steps during which the robot occupies the interior of any group convex hull — providing finer-grained social compliance measurement than terminal-state metrics alone.

\textbf{Contribution 2 — TAGA: Tangent Action for Group Avoidance.}
We present TAGA, a modular group-awareness layer that augments any existing navigation policy — reactive or learned — without retraining. Given group positions and boundaries, TAGA computes tangent trajectories around group convex hulls and coordinates group-level avoidance with individual collision prevention through a hierarchical safety controller. TAGA activates only when a group is detected in the robot's path, returning full control to the base policy otherwise, and requires no changes to the underlying policy weights or training procedure.

Evaluating all methods on our benchmark under both pedestrian dynamics models reveals a consistent empirical pattern we call the \textit{reactive-learning asymmetry}: TAGA provides the largest social compliance gains for reactive baselines (ORCA, Social Force) that carry no internal model of group dynamics. The benefit diminishes for learning-based policies that have implicitly absorbed group-routing behavior through training, and degrades for policies whose training objective explicitly rewards group avoidance. This asymmetry provides actionable guidance: deploy TAGA when retraining is not feasible; use end-to-end group-reward training when a full retraining pipeline is available.

Our specific contributions are:
\textbf{(1) Empirically grounded benchmark} — a five-phase crowd simulation environment (Weidmann speed heterogeneity, Moussa\"{i}d group speed coupling, Kendon F-formations, Helbing leader-follower dynamics, convex-hull group boundaries) evaluated under both ORCA and Social Force pedestrian dynamics.
\textbf{(2) Group Crossing Rate (GCR)} — a continuous, non-terminal metric measuring the fraction of steps the robot spends inside any group convex hull.
\textbf{(3) TAGA} — a modular, training-free group-avoidance layer that augments any existing navigation policy via tangent-path maneuvers with individual-safety override.
\textbf{(4) Reactive-learning asymmetry} — a cross-method empirical finding showing when modular group-awareness adds value and when end-to-end training is preferable.

\section{BACKGROUND AND RELATED WORK}

\subsection{Socially Compliant Robot Navigation}

Enabling mobile robots to navigate safely among pedestrians has been a central challenge in human-robot interaction.
Classical reactive methods such as ORCA~\cite{orca} and the Social Force Model (SFM)~\cite{helbing1995social} established the foundation for collision-free navigation by computing velocity obstacles or attractive-repulsive force fields, respectively.
While computationally efficient, these methods treat every pedestrian as an independent agent with symmetric dynamics, leading to socially inappropriate behaviors such as cutting through conversational clusters or failing to yield to cohesive walking groups~\cite{ferrer2013robot}.
Extensions of SFM have introduced group coherence terms~\cite{zanlungo2011social,moussaid2010walking}, yet their reliance on hand-tuned force parameters limits transferability across environment densities and group types.

Learning-based methods have significantly improved social compliance by modeling crowd dynamics from data.
Inverse reinforcement learning (IRL) approaches~\cite{kretzschmar2016socially} recover latent social cost functions from expert demonstrations, achieving natural-looking trajectories in pedestrian corridors.
Deep reinforcement learning (DRL) extended this capacity to reactive, high-dimensional settings: Chen et al.~\cite{chen2017decentralized} formulated crowd navigation as a pairwise value-network problem (CADRL), later scaled to multi-agent scenarios via social attention~\cite{chen2019crowd}.
Structural-RNN methods~\cite{liu2020decentralized} encode spatio-temporal interaction graphs, capturing richer relational context between the robot and nearby pedestrians, while attention-based intention prediction~\cite{liu2023intention} further improves anticipation of pedestrian goal changes.
Despite their advances, these methods model pedestrian interactions exclusively at the individual level and carry no representation of cohesive group structure.
Because terminal collision metrics record only physical contact events, individual-level policies can achieve strong success rates while routinely penetrating group boundaries undetected --- a gap that motivates the continuous GCR metric we introduce.
This omission is consequential precisely because realistic crowds are composed predominantly of groups rather than isolated individuals~\cite{moussaid2010walking}.

\subsection{Group Detection and Representation}

Detecting and representing human groups is a prerequisite for group-aware navigation.
Static conversational groups---``F-formations'' in proxemics terminology~\cite{kendon1990conducting}---consist of individuals arranged in vis-\`{a}-vis, L-shape, side-by-side, or circular configurations with a shared o-space at the center.
Recognizing such formations has been studied using geometric clustering~\cite{kato2015may} and density-based methods~\cite{ester1996density}.
Vega-Magro et al.\ employed DBSCAN to cluster pedestrians into groups and adapted a navigation architecture to prevent the robot from entering group spaces, but their approach was restricted to static formations and could not handle the dynamic splitting and merging of groups that characterizes real pedestrian traffic~\cite{lu2025garn}.
For dynamic groups, leader-follower models~\cite{moussaid2010walking} capture the cohesion and locomotion of moving clusters, while trajectory prediction networks~\cite{alahi2016social,gupta2018social,9664278} have been used to infer shared group goals from observed motion history.

A key design choice is how the group boundary is represented to the robot.
Zone-based methods define a fixed-radius repulsive zone around each group centroid~\cite{schmuck2025sang}, which is simple but fails to capture elongated or irregular group shapes.
Convex hull representations~\cite{group-aware,lu2025garn} more faithfully enclose the actual footprint of a group regardless of its formation, and have been adopted in recent navigation policies as the basis for group-intrusion reward signals.
Our benchmark environment uses convex hulls to delineate group boundaries (Phase~E), and we adopt the same representation for TAGA's tangent-action computation to ensure consistent and precise group avoidance.

\subsection{Group-Aware Navigation Policies}

Several DRL-based navigation policies have incorporated explicit group modeling.
Katyal et al.~\cite{group-aware} extended the CrowdNav simulation framework to include dynamic group formation and trained a PPO-based policy with a group-aware reward that penalizes convex hull intrusions.
Their key finding is that group awareness during training leads to significantly fewer social norm violations---even when group membership labels are withheld at inference time---demonstrating that group structure can be internalized into the policy's learned representation.
However, their model assumes groups move in the same direction, limiting its application to static or mixed-type group populations, and the policy must be retrained from scratch for any new environment configuration.

SANG (Schmuck \& Celiktutan~\cite{schmuck2025sang}) addresses navigation between stationary conversational groups using an Advantage Actor-Critic (A2C) algorithm with a real-to-sim simulation framework capable of mapping real-world group datasets to 3D environments.
SANG encodes each pedestrian individually with their group label, rather than abstracting each group as a single convex blob, which the authors argue is more faithful to how humans perceive group social entities.
A Navigation Turing Test is introduced as an evaluation protocol, and SANG is judged more human-like 43\% of the time versus 38\% for the prior state-of-the-art.
However, SANG's scenario focuses exclusively on static conversational groups, and the method has not been demonstrated on dynamic groups or on the heterogeneous crowd compositions---combining static F-formations, leader-follower groups, and individual pedestrians---that characterize real public spaces.

GNav (Dang et al.~\cite{dang2026gnav}) employs a spatial-temporal interaction graph to explicitly model group interactions within a DRL framework, demonstrating improved success rates and reduced group intersection rates over individual-level baselines in simulated settings with mixed group-individual compositions.

The most recent and complete group-aware DRL method prior to our work is GARN (Lu et al.~\cite{lu2025garn}), which proposes a Spatio-Temporal Graph Attention Network (STGAN) that encodes obstacle uni-actions, pairwise pedestrian interactions, and group-level interactions jointly in both spatial and temporal domains.
GARN introduces a three-component group-related reward---penalizing group intrusion, incentivizing overtaking, and promoting cooperative passing---and trains end-to-end with model-free DRL using Social Force pedestrian dynamics.
Both simulation and real-world experiments demonstrate that GARN outperforms individual-level baselines across navigation efficiency, group-awareness, and social compliance metrics.
No official code has been released for GARN; we independently reimplement it from the paper description for direct head-to-head evaluation.

Two limitations are shared by the learning-based group-aware methods above.
First, group awareness is embedded inside a monolithic trained policy: adopting any of these approaches for a robot with an existing navigation stack requires full retraining from scratch.
Second, none of these methods has been evaluated under multiple pedestrian dynamics models, leaving open the question of whether strong performance generalizes across crowd behavior assumptions or is specific to the training distribution.
TAGA addresses the first limitation directly --- it is a training-free module that attaches to any existing policy without modifying it.
Our benchmark addresses the second by evaluating all methods under both ORCA and Social Force pedestrian dynamics.

Table~\ref{table:comparison} summarizes the key distinctions between TAGA and related group-aware navigation methods.

\begin{table}[t]
\caption{Comparison of group-aware navigation methods.
  \ding{51}~=~yes; \ding{55}~=~no.
  TF~=~training-free; PI~=~plug-in to any policy;
  SG~=~static groups; DG~=~dynamic groups; MD~=~multi-dynamics eval.}
\label{table:comparison}
\centering
\setlength{\tabcolsep}{5pt}
\renewcommand{\arraystretch}{1.05}
\begin{tabular}{l ccccc}
\toprule
\textbf{Method} & \textbf{TF} & \textbf{PI} & \textbf{SG} & \textbf{DG} & \textbf{MD} \\
\midrule
SANG~\cite{schmuck2025sang}       & \ding{55} & \ding{55} & \ding{51} & \ding{55} & \ding{55} \\
GNav~\cite{dang2026gnav}          & \ding{55} & \ding{55} & \ding{51} & \ding{51} & \ding{55} \\
Katyal et al.~\cite{group-aware}  & \ding{55} & \ding{55} & \ding{51} & \ding{51} & \ding{55} \\
GARN~\cite{lu2025garn}            & \ding{55} & \ding{55} & \ding{51} & \ding{51} & \ding{55} \\
\midrule
\textbf{TAGA (ours)}              & \ding{51} & \ding{51} & \ding{51} & \ding{51} & \ding{51} \\
\bottomrule
\end{tabular}
\end{table}

\subsection{Evaluation Metrics and Benchmarking}

Standard evaluation uses SR, CR, and TR for terminal outcomes~\cite{gao2022evaluation,mavrogiannis2023core}.
Prior group-aware work~\cite{group-aware} treats group intrusion as a terminal failure equivalent to collision, conflating two qualitatively different events and preventing continuous measurement of how deeply the robot intrudes.
We introduce the Group Crossing Rate (GCR) as a continuous, non-terminal metric measuring the fraction of timesteps the robot occupies any group's convex hull interior, complementing SR+CR+TR and providing frame-level social compliance assessment (Section~\ref{sec:benchmark}).

\section{REALISTIC CROWD SIMULATION BENCHMARK}
\label{sec:benchmark}

Existing crowd navigation benchmarks simulate pedestrians as identical agents moving at a uniform preferred speed with no social grouping structure~\cite{chen2017decentralized, liu2020decentralized}. This simplification is empirically unrepresentative: real pedestrian traffic exhibits heterogeneous individual speeds~\cite{weidmann1992social}, and up to 70\% of people in typical public spaces belong to a social group~\cite{moussaid2010walking}. Groups introduce spatial formations, collective motion patterns, and implicit territorial boundaries that fundamentally change the navigation challenge.

We extend the simulation framework of Liu et al.~\cite{liu2023intention} with five empirically grounded modeling phases. Critically, each modeling decision is calibrated to published measurements of real human walking behavior — not heuristically designed — drawing from Weidmann's pedestrian tracking data~\cite{weidmann1992social}, Moussa\"{i}d et al.'s field measurements of walking groups~\cite{moussaid2010walking}, Kendon's sociological studies of conversational formations~\cite{kendon1990conducting}, and Helbing \& Moln\'{a}r's empirical leader-follower dynamics~\cite{helbing1995social}. It is this grounding in real pedestrian behavior data that makes our realism claim substantive: the speed distributions, group compaction factors, formation radii, and locomotion patterns in the simulator match measurements taken from actual human crowds. Each phase is independently controllable so that the benchmark can reproduce both legacy individual-only environments (phases disabled) and our full realistic setting (all phases on). The arena is a circle of radius 8.5~m containing 20 humans: 3 groups of 3--4 members each (2 placed along the robot's direct path) and the remainder navigating independently.

\begin{table}[t]
\caption{Five benchmark phases (independently toggleable).}
\label{table:phases}
\centering
\setlength{\tabcolsep}{4pt}
\renewcommand{\arraystretch}{1.08}
\begin{tabular}{c p{3.6cm} l}
\toprule
\textbf{Ph.} & \textbf{Modeling decision} & \textbf{Source} \\
\midrule
A & $v_{\mathrm{pref}} \sim \mathcal{N}(1.34, 0.26)$ m/s & \cite{weidmann1992social} \\
B & Group speed $= 0.85 \times \min_i v_{\mathrm{pref},i}$ & \cite{moussaid2010walking} \\
C & Static F-formations (4 canonical shapes) & \cite{kendon1990conducting} \\
D & Leader (ORCA) + staggered followers & \cite{helbing1995social} \\
E & Per-step convex hull boundary & \cite{orourke1998comp} \\
\bottomrule
\end{tabular}
\end{table}

Three group behavioral types are randomly assigned per episode: \textit{static\_f} (stationary F-formation), \textit{dynamic\_lf} (leader-follower, Phase~D), and \textit{dynamic\_free} (independent ORCA members, Phase~B). The benchmark is evaluated under two pedestrian dynamics: \textbf{ORCA}~\cite{orca} (smooth, reciprocal avoidance) and \textbf{Social Force}~\cite{helbing1995social} (oscillatory, closer passing distances), allowing assessment of cross-dynamics generalization.

\subsection{Group Crossing Rate (GCR)}

Standard terminal-state metrics (SR, CR, TR) cannot distinguish a robot that narrowly clips a group boundary from one that traverses group space for extended periods. We introduce the \emph{Group Crossing Rate} (GCR) as a continuous, non-terminal social compliance metric:
\begin{equation}
\mathrm{GCR} = \frac{1}{T} \sum_{t=1}^{T} \mathbb{I}\!\left(\exists\,j : r_t \in \mathcal{H}_j^{(t)}\right) \times 100\%,
\end{equation}
where $T$ is the total control steps in the episode, $\mathcal{H}_j^{(t)}$ is the convex hull polygon of group $j$ rebuilt from member positions at timestep $t$, and the indicator is 1 if the robot position $r_t$ lies inside any group hull. GCR is evaluated at every step regardless of the terminal outcome, providing frame-level resolution of social compliance that complements SR, CR, and TR (which sum to 1 independently of GCR).

\section{TANGENT ACTION MODEL FOR GROUP AVOIDANCE}
\label{sec:methods}

\subsection{Problem Formulation and Group Detection}

Our objective is to enable a robot to navigate to its goal while respecting both individual humans and group formations in dynamic environments.
TAGA operates as a conditional layer: when a blocking group is detected it produces a brief tangent avoidance nudge; otherwise the base policy retains full control.

Each group $g_j$ is represented by its centroid $C_j$, bounding radius $R_j$ (minimum enclosing circle of its convex hull), member index set $G_j$, and mean velocity $\mathbf{v}_j = \frac{1}{|G_j|}\sum_{i \in G_j} \mathbf{v}_i$.
In simulation we use ground-truth group membership, centroid, and hull geometry provided by the simulator, which isolates the avoidance strategy from detection noise and establishes an upper bound on achievable performance.
For real-world deployment, a DBSCAN-based detector with proximity threshold $\epsilon = 1.5$\,m and velocity-alignment threshold $v_{\mathrm{thresh}}$ provides a drop-in replacement:
\begin{equation}
G_j = \{h_i \mid d(h_i, h_k) < \epsilon,\; |\mathbf{v}_i - \mathbf{v}_k| < v_{\mathrm{thresh}},\; \forall\, h_k \in G_j\}
\end{equation}

The robot observes $s_t = \{w_t,\, u_1^t,\!\ldots,u_n^t\} \cup \{g_1^t,\!\ldots,g_m^t\}$, where $w_t$ encodes robot pose and velocity, $\{u_i^t\}$ are individual human states, and $\{g_j^t\}$ are detected group descriptors.

\begin{figure}[!t]
\centering
\includegraphics[width=3.5in]{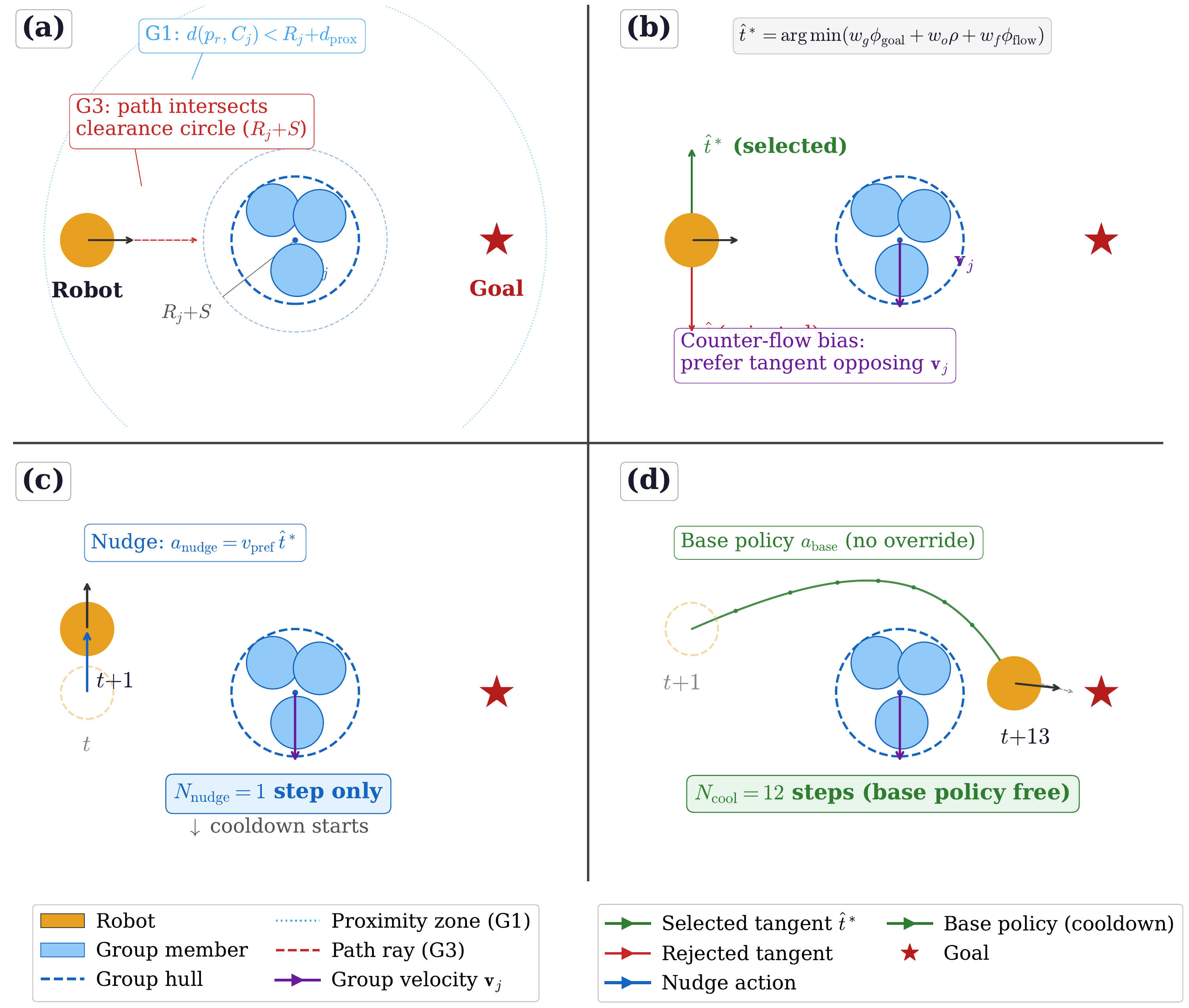}
\caption{TAGA nudge design. When the robot enters the proximity zone (group boundary + 2\,m), it checks whether the current velocity intersects the clearance circle. If so, it selects the lower-cost tangent direction and applies a single-step nudge at $v_{\mathrm{pref}}$, followed by a cooldown window where the base policy runs freely. Group velocity $\mathbf{v}_j$ biases the selection toward the counter-flow side.}
\label{fig:taga-process}
\end{figure}

\subsection{Nudge-Based Group Avoidance}

Prior tangent-based approaches apply a continuous blended override whenever a group is within switching distance, which causes the correction to fire for 10--30 steps per encounter.
For classical policies such as ORCA this is manageable---ORCA recomputes from scratch each step and recovers immediately from any detour---but for learning-based policies with recurrent hidden states (e.g., DS-RNN, Intention-RL), sustained overrides accumulate out-of-distribution transitions that degrade performance.

We address this with a \emph{nudge-based} design: TAGA fires a single directed action step and then withdraws for a fixed cooldown, giving the base policy space to re-route.
The full per-step logic is:

\begin{equation}
\pi(s_t) = \begin{cases}
a_{\mathrm{nudge}}  & \text{fresh trigger at } t \\
a_{\mathrm{nudge}}  & \text{continuation, step } 1 \leq k < N_{\mathrm{nudge}} \\
a_{\mathrm{base}}   & \text{cooldown, step } N_{\mathrm{nudge}} \leq k < N_{\mathrm{nudge}} + N_{\mathrm{cool}} \\
a_{\mathrm{base}}   & \text{no blocking group, goal priority, or path clear}
\end{cases}
\label{eq:nudge-logic}
\end{equation}

where $N_{\mathrm{nudge}} = 1$ (single-step burst by default) and $N_{\mathrm{cool}} = 12$ (three seconds at $\Delta t = 0.25$\,s).
A fresh trigger is issued only when all four activation gates described below pass simultaneously.

\subsection{Activation Gates}

\textbf{(G1) Proximity gate.}
TAGA is considered for group $g_j$ only when the robot is within a proximity threshold of the group boundary:
\begin{equation}
d(p_r, C_j) < R_j + d_{\mathrm{prox}}, \quad d_{\mathrm{prox}} = 2.0\,\text{m}
\end{equation}
This prevents premature activation when the base policy would naturally avoid the group before reaching it.

\textbf{(G2) Forward-direction gate.}
The group must lie in the robot's forward half-space relative to the goal:
\begin{equation}
(\mathbf{p}_{\mathrm{goal}} - \mathbf{p}_r) \cdot (C_j - \mathbf{p}_r) > 0
\quad \text{and} \quad
d(p_r, \mathrm{goal}) > d(C_j, \mathrm{goal})
\end{equation}
Groups already behind the robot or closer to the goal than the robot are ignored.

\textbf{(G3) Path-intersection gate.}
TAGA activates only if the robot's current velocity direction $\hat{a}_{\mathrm{base}}$ will geometrically enter the clearance circle of radius $R_j + S$ (safety margin $S = 0.5$\,m) within lookahead distance $L = 5.0$\,m.
Formally, let $\mathbf{b} = \hat{a}_{\mathrm{base}}$, $\mathbf{q} = p_r - C_j$:
\begin{equation}
\begin{split}
b &= \mathbf{b}^\top\mathbf{q}, \quad c = |\mathbf{q}|^2 - (R_j+S)^2 \\
\text{activate iff} &\quad b^2 - c \geq 0 \;\;\text{and}\;\;
  t_2 = -b + \sqrt{b^2 - c} \geq 0
\end{split}
\end{equation}
with $t_1 \leq L$.
If the current action already avoids the group the gate suppresses the nudge, preventing interference on steps where the base policy is already navigating correctly.

\textbf{(G4) Individual-safety gate.}
Before committing to a nudge direction, TAGA casts a capsule of width $r_r + r_h + 0.2$\,m along the proposed velocity for a lookahead of 2.0\,m.
Pedestrian $u_i$ (with $i \notin G_j$) blocks the path if its projection onto the capsule axis satisfies $0 < d_{\mathrm{proj}} \leq 2.0$ and its perpendicular offset is within the capsule width.
If the preferred tangent direction is blocked, the alternate direction is tried.
If both are blocked, the nudge is skipped entirely and $a_{\mathrm{base}}$ is used.
This ensures individual interactions remain the responsibility of the base policy.

\subsection{Tangent Direction with Counter-Flow Bias}

For each blocking group $g_j$, the two perpendicular unit directions from the robot toward the group centroid are:
\begin{equation}
\hat{t}_{\mathrm{cw}}  = \frac{[-q_y,\; q_x]^\top}{|q|}, \quad
\hat{t}_{\mathrm{ccw}} = \frac{[q_y,\; -q_x]^\top}{|q|}, \quad
\mathbf{q} = C_j - p_r
\end{equation}

The direction with lower total cost is selected:
\begin{equation}
\hat{t}^* = \arg\min_{\hat{t} \in \{\hat{t}_{\mathrm{cw}},\, \hat{t}_{\mathrm{ccw}}\}}
\bigl(w_g\,\phi_{\mathrm{goal}}(\hat{t}) + w_o\,\rho(\hat{t}) + w_f\,\phi_{\mathrm{flow}}(\hat{t})\bigr)
\label{eq:cost}
\end{equation}

where $\phi_{\mathrm{goal}}(\hat{t}) = (1 - \hat{t} \cdot \hat{g}) / 2$ is the normalized angular deviation from the goal direction $\hat{g}$,
$\rho(\hat{t})$ is a forward-cone obstacle density score summing $1/(d{+}0.1)$ over all humans and groups within a $60^\circ$ half-angle cone at range up to 5\,m,
and $\phi_{\mathrm{flow}}(\hat{t}) = \hat{v}_j \cdot \hat{t}$ is the \emph{counter-flow bias}: it penalizes directions aligned with the group's own travel direction $\hat{v}_j = \mathbf{v}_j / |\mathbf{v}_j|$, so the robot naturally moves against the group's flow and the two agents separate rather than travel in parallel.
The weights are $w_g = 0.4$, $w_o = 0.6$, $w_f = 0.4$ (applied only when $|\mathbf{v}_j| > 0.05$\,m/s).

The nudge action is then:
\begin{equation}
a_{\mathrm{nudge}} = v_{\mathrm{pref}}\,\hat{t}^*
\label{eq:nudge-action}
\end{equation}

For classical policies (ORCA, Social Force) the nudge overrides the internal action computed at that step via the \texttt{taga\_action\_override} hook, preventing a double-computation of the internal velocity-obstacle solver.

\subsection{Integration and Switching Mechanism}

TAGA requires no modifications to the base policy weights or training procedure and attaches to any navigation policy at test time. The full per-step logic follows directly from the nudge-based design of Fig.~\ref{fig:taga-process}: evaluate gates G1--G3 in order, compute $\hat{t}^*$ via Eq.~\eqref{eq:cost}, apply G4, fire a single nudge step, then release control for $N_{\mathrm{cool}}$ steps.

\section{EVALUATION SETUP}
\label{sec:eval}

\begin{figure}[t]
    \centering
    \includegraphics[width=\columnwidth]{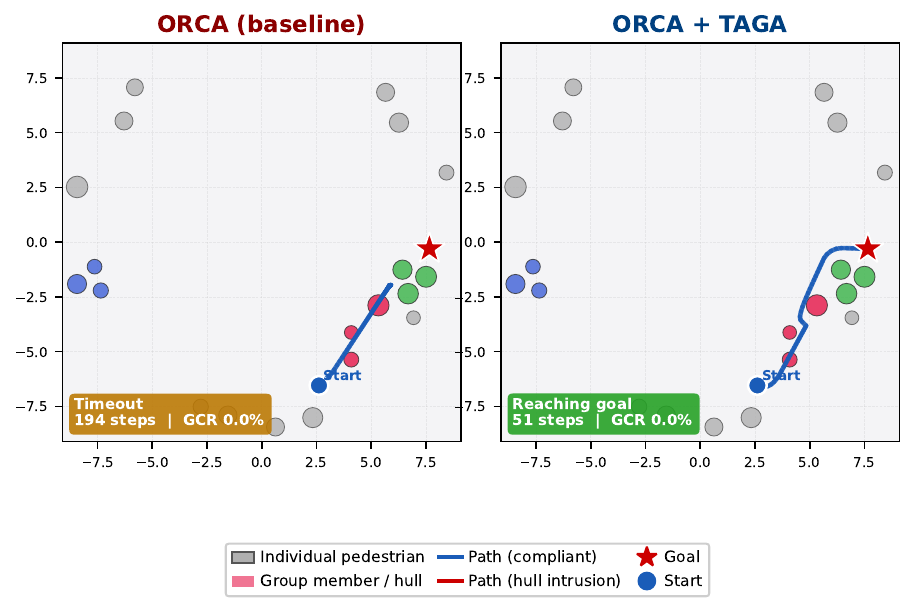}
    \vspace{-14pt}
    \caption{Trajectory comparison on a single episode (seed~0).
    Each panel shows the \textbf{complete robot path} overlaid on the initial environment snapshot.
    \textcolor[HTML]{1a5cb8}{Blue}: compliant path outside group hulls.
    \textcolor[HTML]{cc0000}{Red}: hull-intrusion segment.
    \textbf{Left (ORCA alone):} two groups fill the velocity-obstacle feasible set symmetrically; the robot oscillates and times out after 194 steps.
    \textbf{Right (ORCA+TAGA):} a single tangent nudge displaces the robot laterally, breaking the deadlock; the robot reaches the goal in 51 steps with zero hull intrusion.}
    \vspace{-8pt}
    \label{fig:sim-com}
\end{figure}

\subsection{Evaluation Protocol}

We evaluate all methods under two complementary settings.

\textbf{Individual-only environment.}
All 20 pedestrians navigate independently (no groups) under either ORCA or Social Force dynamics. This establishes baseline navigation competence before group complexity is introduced; GCR is not applicable and is omitted.

\textbf{Group environment.}
The full five-phase benchmark (Section~\ref{sec:benchmark}) is active. Each navigation method is evaluated both as a standalone policy and augmented with TAGA. This quantifies (i)~how much each base policy degrades in social compliance when groups are present, and (ii)~how much TAGA recovers compliance without retraining the base policy.

Episode parameters: arena radius 8.5~m, 20 humans (3 groups $\times$ 3--4 members, 2 on-path; remaining individuals), time limit 48.25~s (193 steps at $\Delta t = 0.25$~s). Episodes reaching the time limit are recorded as timeouts. The robot is holonomic with a 5~m sensing range and preferred speed 1~m/s. All methods are evaluated on 500 randomly generated episodes with fixed seeds for reproducibility.

\subsection{Baselines and Implementation}

We evaluate three navigation policies as baselines:

\textbf{ORCA}~\cite{orca}: classical reactive method using reciprocal velocity obstacles. No learning component.

\textbf{Social Force (SF)}~\cite{helbing1995social}: classical reactive method using attractive goal forces and repulsive inter-agent forces. No learning component.

\textbf{DS-RNN}~\cite{liu2020decentralized}: decentralised structural RNN that models spatial--temporal interactions between the robot and nearby humans via graph-based recurrent networks.

\textbf{Intention-RL}~\cite{liu2023intention}: attention-based DRL with a GST predictor~\cite{9664278} for anticipating pedestrian goal changes; the strongest individual-level learned baseline in our benchmark.

TAGA is applied on top of each baseline as an additional evaluation condition in the group environment. We also include \textbf{GARN}~\cite{lu2025garn} as a standalone reference: the most recent published group-aware DRL method, evaluated without TAGA since its training objective already internalizes group avoidance. No official code has been released for GARN; we re-implement it from the paper description.

Key TAGA parameters: safety margin $S = 0.3$~m, emergency distance $d_{\mathrm{crit}} = 0.3$~m, safety distance $d_{\mathrm{safe}} = 0.5$~m, individual proximity lookahead $d_{\mathrm{look}} = 2\,\Delta t\,v_{\mathrm{pref}} = 0.5$~m.

\subsection{Metrics}

\textbf{Terminal state:} Success Rate (SR), Collision Rate (CR), Timeout Rate (TR), where SR\,+\,CR\,+\,TR\,$=$\,1.

\textbf{GCR:} fraction of navigation steps the robot spends inside any group boundary (Section~\ref{sec:benchmark}); lower is more socially compliant.

\textbf{Efficiency:} Navigation Time (NT, seconds) and Path Length (PL, metres) are reported for successful episodes only.

\section{RESULTS AND DISCUSSION}
\label{sec:results}

Table~\ref{table:group} is the central result: each base method versus its TAGA-augmented variant in the full group environment under both pedestrian dynamics models. Figure~\ref{fig:sim-com} provides qualitative context. In the individual-only baseline (no groups, 100 episodes), SR ranged from 0.54--0.76 under ORCA pedestrians and 0.34--0.86 under Social Force pedestrians, confirming each method handles standard crowd navigation before group complexity is introduced; the SF navigation policy is notably harder, dropping to SR\,=\,0.34.

\subsection{Group Environment}

\begin{table*}[t]
\caption{Navigation in the group environment (five-phase benchmark, Section~\ref{sec:benchmark}). Each base method vs.\ its TAGA-augmented variant under ORCA and Social Force pedestrian dynamics. GCR (\%) = fraction of steps inside any group convex hull (lower is better). Averaged over 500 episodes.}
\label{table:group}
\centering
\setlength{\tabcolsep}{4pt}
\begin{tabular}{l ccccc | ccccc}
\toprule
& \multicolumn{5}{c|}{\textbf{ORCA Pedestrians}} & \multicolumn{5}{c}{\textbf{Social Force Pedestrians}} \\
\textbf{Method}
  & \textbf{SR$\uparrow$} & \textbf{CR$\downarrow$} & \textbf{TR$\downarrow$} & \textbf{GCR$\downarrow$} & \textbf{NT~(s)}
  & \textbf{SR$\uparrow$} & \textbf{CR$\downarrow$} & \textbf{TR$\downarrow$} & \textbf{GCR$\downarrow$} & \textbf{NT~(s)} \\
\midrule
ORCA           & 0.73 & 0.21 & 0.06 & 0.52 & 19.03
               & 0.87 & 0.08 & 0.05 & 1.01 & 20.03 \\
ORCA\,+\,TAGA  & \textbf{0.81} & \textbf{0.14} & 0.05 & \textbf{0.30} & 19.63
               & \textbf{0.88} & \textbf{0.06} & 0.06 & \textbf{0.84} & 19.23 \\
\midrule
SF             & 0.40 & 0.39 & 0.21 & 0.07 & 21.99
               & 0.47 & 0.48 & 0.05 & 0.25 & 23.03 \\
SF\,+\,TAGA    & \textbf{0.47} & \textbf{0.38} & \textbf{0.15} & \textbf{0.05} & 24.55
               & 0.47 & 0.48 & 0.05 & 0.43 & 21.77 \\
\midrule
DS-RNN~\cite{liu2020decentralized}
               & 0.65 & 0.25 & 0.10 & 0.54 & 20.98
               & 0.68 & 0.23 & 0.09 & 0.78 & 19.98 \\
DS-RNN\,+\,TAGA & \textbf{0.68} & 0.26 & \textbf{0.06} & \textbf{0.32} & 21.13
               & \textbf{0.69} & \textbf{0.22} & 0.09 & \textbf{0.45} & 20.12 \\
\midrule
Intention-RL~\cite{liu2023intention}
               & \textbf{0.83} & \textbf{0.15} & 0.02 & 0.40 & 15.50
               & \textbf{0.85} & \textbf{0.15} & 0.00 & 0.34 & 15.87 \\
IRL\,+\,TAGA   & 0.82 & 0.17 & \textbf{0.01} & \textbf{0.36} & 15.76
               & \textbf{0.85} & \textbf{0.15} & 0.00 & \textbf{0.06} & 15.75 \\
\bottomrule
\end{tabular}
\end{table*}

Table~\ref{table:group} and Figure~\ref{fig:sim-com} reveal three consistent patterns across both pedestrian dynamics conditions.

\textbf{Pattern 1 — TAGA improves reactive baselines.}
For ORCA navigation, TAGA increases SR by 8\,pp under ORCA pedestrians (0.73\,$\rightarrow$\,0.81) and 1\,pp under Social Force pedestrians (0.87\,$\rightarrow$\,0.88), while reducing GCR in both conditions (0.52\,$\rightarrow$\,0.30\% and 1.01\,$\rightarrow$\,0.84\%).
The larger SR gain under ORCA pedestrians reflects harder navigation conditions where group avoidance decisions are more consequential; the GCR reductions confirm TAGA successfully enforces the social compliance objective it was designed for.

\textbf{Pattern 2 — Social Force navigation is a mixed case.}
The SF navigation model already struggles in the group environment (SR\,=\,0.40 under ORCA pedestrians), a consequence of the oscillatory behavior inherent to force-field planning near obstacles.
TAGA recovers 7\,pp SR under ORCA pedestrians and reduces GCR (0.07\,$\rightarrow$\,0.05\%) while maintaining parity under SF pedestrians.
However, GCR \emph{increases} under SF pedestrians (0.25\,$\rightarrow$\,0.43\%): the tighter group spacing produced by SF pedestrian dynamics occasionally routes the tangent detour into an adjacent group boundary.
Activating TAGA's group avoidance only for static or slow-moving groups — and deferring to the base policy for fast-moving ones — is a direct direction for future improvement.

\textbf{Pattern 3 — TAGA has near-neutral impact on learned policies.}
For Intention-RL, adding TAGA causes only a 1\,pp SR drop under ORCA pedestrians (0.83\,$\rightarrow$\,0.82) and no change under Social Force pedestrians (0.85\,$\rightarrow$\,0.85).
GCR is reduced in both conditions (0.40\,$\rightarrow$\,0.36\% and 0.34\,$\rightarrow$\,0.06\%), confirming TAGA still enforces social compliance.
Trained on thousands of episodes in the group environment, Intention-RL has implicitly absorbed group-routing heuristics into its value function; the brief single-step nudge (Section~\ref{sec:methods}) does not disrupt this sufficiently to damage navigation success.
The nudge-based design --- one override step followed by a 12-step free-recovery window --- is the key factor: earlier continuous-override designs cost 10\,pp SR on this class of policy.

\textbf{Summary — reactive-learning asymmetry.}
The three patterns define a principled asymmetry: TAGA adds the most value where the base policy has the least group intelligence (ORCA: +8\,pp SR, GCR halved; see Fig.~\ref{fig:sim-com} where TAGA resolves an ORCA deadlock by displacing the robot out of a symmetric velocity-obstacle trap), moderate benefit for DS-RNN (+3\,pp SR, GCR $-41$\%), and near-zero cost for strongly trained policies (Intention-RL: $-$1\,pp SR).
The key mechanism: ORCA and Social Force recompute from scratch each step and recover immediately from any detour, while recurrent policies carry hidden state that a brief nudge does not corrupt sufficiently to matter.
The practical implication: TAGA is safe to deploy on any stack --- the worst observed cost is 1\,pp SR --- and provides the largest gains for classical reactive baselines.

\section{CONCLUSION AND FUTURE WORK}

This paper presented two peer-level contributions for socially-compliant robot navigation in realistic crowds. The \textbf{realistic crowd simulation benchmark} (Section~\ref{sec:benchmark}) extends existing simulation frameworks with five empirically grounded phases — individual speed heterogeneity~\cite{weidmann1992social}, group speed coupling~\cite{moussaid2010walking}, F-formation static groups~\cite{kendon1990conducting}, leader-follower moving groups~\cite{helbing1995social}, and convex-hull group boundaries — evaluated under both ORCA and Social Force pedestrian dynamics. The \textbf{Group Crossing Rate (GCR)} metric provides continuous frame-level assessment of social compliance that terminal-state metrics cannot capture. \textbf{TAGA} (Section~\ref{sec:methods}) is a modular group-avoidance layer that augments any existing navigation policy with tangent-path maneuvers around group convex hulls and a hierarchical safety controller with individual proximity override, without retraining.

Our central finding is a \textit{reactive-learning asymmetry}: TAGA provides the largest gains for classical reactive baselines (ORCA, Social Force) that have no built-in model of group dynamics, while adding zero or negative value to learned policies that have already internalized group-routing behavior. This provides actionable guidance: deploy TAGA when augmenting a classical navigation stack; use end-to-end group reward training when retraining is feasible.

\textbf{Limitations.} TAGA's reactive geometry cannot anticipate group movement, leading to unnecessary detours when groups are moving away from the robot. The GCR increase observed with SF pedestrians under Social Force navigation (Section~\ref{sec:results}) reveals a specific failure mode: tangent detours can route the robot into adjacent groups when group spacing is tight. A dynamic-vs-static group classifier that activates TAGA only for static or slow groups would directly address this.

\textbf{Sim-to-real transfer.}
All evaluations are in simulation; real-world deployment is the critical next step. A promising path toward closing the sim-to-real gap is to embed authentic human motion trajectories from real pedestrian datasets~\cite{lerner2007crowds, pellegrini2009you} as replay agents in the simulation. These data-driven agents would expose navigation policies to the statistical distribution of real crowd behavior — heterogeneous social norms, asymmetric reaction times, culture-specific proxemics — that analytic models approximate imperfectly. Combining this data-driven crowd environment with domain randomization over crowd density and group size would provide a principled pipeline from simulation training to physical robot deployment. We view this as the most impactful direction toward socially-compliant navigation that generalizes beyond controlled benchmarks.

\addtolength{\textheight}{-10cm}


\bibliographystyle{IEEEtran}
\bibliography{citations}

\end{document}